\documentclass{isprs}
\usepackage{subfigure}
\usepackage{setspace}
\usepackage{geometry} % added 27-02-2014 Markus Englich
\usepackage{enumerate}
\usepackage{natbib}
\usepackage{multirow}  % 多行合并的宏包
\usepackage{stfloats}
 \usepackage{url}

\begin{document}

\title{Directionally Constrained Fully Convolutional Neural Network for Airborne LiDAR Point Cloud Classification}

\author{
Congcong Wen\textsuperscript{a,b}, Lina Yang\textsuperscript{a,b}, Ling Peng\textsuperscript{a,b}, Xiang Li\textsuperscript{a,b,c}\thanks{Corresponding author. Email: lixiang01@radi.ac.cn.,  %This is useful to know for communication with the appropriate person in cases with more than one author.
% This work was done when Xiang Li is partial at the Institute of Remote Sensing and Digital Earth, Chinese Academy of Sciences.
}, Tianhe Chi\textsuperscript{a,b}
}

\address
{
\textsuperscript{a }Institute of Remote Sensing and Digital Earth, Chinese Academy of Sciences, Beijing, China.\\
\textsuperscript{b }
University of Chinese Academy of Sciences, Beijing, China.\\
\textsuperscript{c }Tandon School of Engineering, New York University, New York, United States.\\
}
%  - wencc@radi.ac.cn
%   - xl1845@nyu.edu
% \commission{VI, }{VI} %This field is optional.
% \workinggroup{VI/4} %This field is optional.

\abstract
{
Point cloud classification plays an important role in a wide range of airborne light detection and ranging (LiDAR) applications, such as topographic mapping, forest monitoring, power line detection, and road detection. However, due to the sensor noise, high redundancy, incompleteness, and complexity of airborne LiDAR systems, point cloud classification is challenging. Traditional point cloud classification methods mostly focus on the development of handcrafted point geometry features and employ machine learning-based classification models to conduct point classification. In recent years, the advances of deep learning models have caused researchers to shift their focus towards machine learning-based models, specifically deep neural networks, to classify airborne LiDAR point clouds. These learning-based methods start by transforming the unstructured 3D point sets to regular 2D representations, such as collections of feature images, and then employ a 2D CNN for point classification. Moreover, these methods usually need to calculate additional local geometry features, such as planarity, sphericity and roughness, to make use of the local structural information in the original 3D space. Nonetheless, the 3D to 2D conversion results in information loss.
In this paper, we proposed a directionally constrained fully convolutional neural network (D-FCN) that can take the original 3D coordinates and LiDAR intensity as input; thus, it can directly apply to unstructured 3D point clouds for semantic labeling. Specifically, we first introduce a novel directionally constrained point convolution (D-Conv) module to extract locally representative features of 3D point sets from the projected 2D receptive fields. To make full use of the orientation information of neighborhood points, the proposed D-Conv module performs convolution in an orientation-aware manner by using a directionally constrained nearest neighborhood search. Then, we designed a multiscale fully convolutional neural network with downsampling and upsampling blocks to enable multiscale point feature learning. The proposed D-FCN model can therefore process input point cloud with arbitrary sizes and directly predict the semantic labels for all the input points in an end-to-end manner.
Without involving additional geometry features as input, the proposed method has demonstrated superior performance on the International Society for Photogrammetry and Remote Sensing (ISPRS) 3D labeling benchmark dataset. The results show that our model has achieved a new state-of-the-art level of performance with an average F1 score of 70.7\%, and it has improved the performance by a large margin on categories with a small number of points (such as powerline, car, and facade).

}


\maketitle

\section{Introduction}\label{Introduction}

The airborne light detection and ranging (LiDAR) system provides a new technical approach for acquiring 3D spatial data. Unlike commonly used satellite images, which mainly provide spectral information in a regular spatial grid, a LiDAR system captures data in 3D point cloud format, which provides reliable depth information and can be used to accurately localize objects and characterize their structure. Among various point cloud recognition tasks, the point cloud classification problem has been an active research area in remote sensing communities for decades \citep{rabbani2006segmentation,vosselman2017contextual} and serves as an essential building block for many applications, including topographic mapping, forest monitoring \citep{axelsson2000generation,mongus2013computationally}, power line detection \citep{andersen2005estimating,solberg2009mapping,zhao2009lidar,ene2017large}, road detection and planning, 3D building reconstruction \citep{kada20093d,yang2017automated}, and others. However, this process is still challenging due to the unstructured nature of 3D point clouds, their limited data resolution and high sensor noise, uneven density distribution, incompleteness and scene complexity \citep{zhu2017robust}.

In general, the point cloud classification task can be divided into two basic steps:  extracting representative point features both locally and globally and classifying each point into pre-defined semantic categories using the learned feature representations.
Early works developed various hand-crafted point descriptors based on the geometric structure of local neighborhood points, e.g., density, curvature, roughness.
In these methods, machine learning algorithms such as the Gaussian Mixture Model \citep{lalonde2005scale,lalonde2006natural}, Support Vector Machine \citep{zhang2013svm}, AdaBoost \citep{lodha2007aerial}, and Random Forest \citep{babahajiani2017urban,chehata2009airborne} are commonly used to achieve point cloud classification based on hand-crafted features. However, these methods estimate the local features of each point independently and generate the label predictions without considering label consistency among neighborhood points. Therefore, the classification results tend to suffer from noise and label inconsistency.

Some studies have attempted to address this problem by incorporating contextual information into the classification models; the representative works include associative Markov networks \citep{munoz2009contextual}, non-associative Markov networks \citep{shapovalov2010nonassociative} and conditional random fields \citep{niemeyer2014contextual,weinmann2015contextual,niemeyer2012conditional}.
Nevertheless, these machine learning-based point cloud classification methods comprehensively employ hand-crafted features to characterize each point in the input point cloud; thus, they have limited generalizability when applied to large-scale wild scenes.

In recent years, one of the most prevalent types of artificial intelligence algorithms (i.e., deep learning methods) has achieved great success in various real-world applications, including image classification, speech recognition, time series prediction, natural language processing and so on \citep{lecun2015deep,chan2015pcanet,hinton2012deep,wen2019novel,collobert2008unified,li2016deep,li2018building}. Following this trend, researchers have shifted their focus towards learning-based approaches and, more specifically, to deep neural networks \citep{qi2016volumetric,su2015multi,qi2017pointnet,qi2017pointnet++} for learning 3D local descriptors.
For example, the authors of \citep{yang2017convolutional,zhao2018classifying} proposed first transforming 3D point sets into regular 2D image representations and then employing a conventional 2D CNN for point classification. Such methods usually need to calculate additional local geometry features, such as planarity, sphericity and roughness, to enrich the 2D representations and further enhance the classification performance. However, the 3D to 2D conversion process results in information loss. More recent works have tried to apply convolution directly to unstructured point sets for classification \citep{yousefhussien2018multi,wang2018deep}; however, these methods also require additional hand-crafted features (i.e., optical image features, local geometric features) as input to ensure the classification accuracy.

In this paper, we propose a novel directionally constrained fully convolutional network (D-FCN) that can be directly applied to unstructured 3D point clouds for airborne LiDAR point cloud classification. The raw 3D coordinates (i.e. X, Y, and Z) and intensity values obtained from the airborne LiDAR data are used as the input of our proposed model. Then, we introduce a novel directionally constrained point convolution (D-Conv) module to extract the locally representative features of 3D point sets. Based on this convolution module, we further design a multiscale fully convolutional neural network with downsampling and upsampling blocks to enable multiscale point feature learning. The resulting network can be trained in an end-to-end manner and can directly predict the semantic labels for all the input points. The main contributions of the proposed method are as follows:

\begin{enumerate}[1.]
\item This paper proposes a directionally constrained fully convolution neural network (D-FCN) that can be directly applied to raw point clouds for semantic labeling. The proposed D-FCN model can process input point cloud with arbitrary sizes and directly predict the semantic labels for all the input points in an end-to-end manner.
\item This paper introduces a novel directionally constrained point convolution (D-Conv) module to extract locally representative features of 3D point sets from the projected 2D receptive fields. Specifically, the proposed D-Conv module performs convolution in an orientation-aware manner by using a directionally constrained nearest neighborhood search.
\item This paper proposes a multiscale fully convolutional neural network with downsampling and upsampling blocks to enable multiscale point feature learning.
\item Without involving additional geometry features as input, the proposed method demonstrates superior performance on the ISPRS 3D labeling benchmark dataset.

\end{enumerate}

The remainder of this paper is organized as follows. In Section 2, we provide a brief review of the airborne LiDAR point cloud classification methods. The proposed directional fully convolutional network (D-FCN) is described in detail in Section 3.
In Section 4, we conduct experiments to verify the classification performance of the proposed method. Finally, the paper is concluded in Section 5.

\section{Related Work}\label{Related Work}

\subsection{Point-based feature classification method}
Traditional point cloud classification methods commonly regard each point in a point set as an independent entity, and classifying these points only depends on the local geometry features. \citeauthor{zhang2013svm} first calculated thirteen features of geometry, radiometry, topology and echo characteristics and then employed the Support Vector Machine (SVM) algorithm to classify airborne LiDAR point clouds of urban scene \citep{zhang2013svm}. \citeauthor{lodha2007aerial} adopted the AdaBoost algorithm to classify airborne LiDAR point clouds into four categories (i.e. road, grass, buildings, and trees) based on five features including height, height variation, normal variation, intensity, and image intensity \citep{lodha2007aerial}. \citeauthor{chehata2009airborne} used the random forest algorithm to achieve the point cloud classification by selecting the most relevant features among the 21 features, which is composed of seventeen multi-echo and four full-waveform LiDAR features
\citep{chehata2009airborne}.

However, the estimation of single-point local features is unstable due to the uneven distribution of point clouds, which can easily lead to classification noises and label inconsistency, especially for complex scenes \citep{weinmann2015semantic}.

\subsection{Context-based features classification method}
To obtain smoother and more accurate classification results, many studies have introduced contextual information for point cloud classification based on single point local features. The associative Markov network (AMN) was presented by \citeauthor{munoz2009contextual} to extract various features comprising the degree of scatter, linearity, planarity, etc.. These features contain multi-level contextual information. Then, contextual interactions were discriminatively considered to conduct 3D LiDAR point cloud classification \citep{munoz2009contextual}. Although the AMN model involving high-order interactions achieved more robust classification results than previous methods, it failed to detect both large and small objects due to oversmoothing, and it may lead to incorrect classifications of large region due to error propagation \citep{shapovalov2010nonassociative}. To address this issue, \citeauthor{shapovalov2010nonassociative} further proposed the non-associative Markov network to overcome the weakness of the AMN model by using dynamic instead of constant pairwise potentials for pairs of different class labels based on 68-dimensional input features, including spectral and directional features, spin images, angular spin images and height distributions. In addition, \citeauthor{niemeyer2012conditional} designed ten features, including amplitude, echo width, normalized echo number, distance to ground and so on, as the model input, and adopted the conditional random field (CRF) approach to classify airborne LiDAR point clouds. This proposed method enables contextual information to be incorporated \citep{niemeyer2012conditional}.

However, these classification methods require manually extracting context features in advance, which complicates the classification task. Moreover, their classification performances degrade when dealing with point clouds scanned from complex scenes \citep{zhao2018classifying}.

\subsection{Deep learning-based classification method}
As deep learning became increasingly prevalent, research on 3D point signature learning has followed the general trend in the machine vision community and shifted towards learning-based approaches for 3D point cloud classification.

In general, deep learning-based point cloud classification methods can be divided into two categories: feature image-based methods and point cloud-based methods.

\subsubsection{Feature image based classification method}~\\
Convolutional neural networks (CNNs) have achieved great success on various 2D image recognition tasks, including scene classification, object detection, semantic segmentation, and many others. Following the success of 2D CNNs, many studies have tried to generate feature images from point clouds and then employ convolutional neural networks to achieve airborne LiDAR point cloud classification \citep{yang2017convolutional,yang2017convolutional,yang2018segmentation,zhao2018classifying}. For example, \citeauthor{yang2017convolutional} proposed an image-generation method that transformed the local geometric features, global geometric features and full-waveform features of each point in the airborne LiDAR point clouds into 2D feature images and then applied conventional 2D convolutional neural networks to classify the feature images \citep{yang2017convolutional}.
\citeauthor{yang2018segmentation} generated feature images of height, intensity, planarity, sphericity and variance of deviation angles across multiple scales and developed a multiscale convolutional neural network that learns the deep features of each point from the above feature images to enable airborne LiDAR point cloud classification. \citep{yang2018segmentation}. \citeauthor{zhao2018classifying} obtained a set of contextual images by calculating LiDAR point set features, including height, intensity and roughness, and then proposed a multiscale convolutional neural network to classify the airborne LiDAR point clouds into several object categories \citep{zhao2018classifying}.

However, these studies all involve generating 2D feature images from 3D point clouds, which can cause spatial information loss, and they still require handcrafted feature engineering before network training. In the feature engineering process, the selected features do not necessarily represent the most critical information of the original point cloud, and the manually defined combinations of different features can have a substantial influence on the final classification results.

\subsubsection{Point cloud-based classification method}~\\
Due to the unordered and unstructured nature of point clouds, early efforts in the 3D point cloud recognition domain mainly attempted to convert 3D point clouds into more familiar representations and then apply conventional 2D/3D CNN to learn point features. Recently, \cite{qi2017pointnet} proposed a deep learning framework called PointNet that can be applied directly to unstructured point clouds and learn representative point signatures from massive input point sets in a data-driven fashion. Experiments with various point cloud recognition tasks, including 3D shape classification, 3D ShapeNet part segmentation, 3D scene semantic segmentation and 3D object detection tasks, demonstrated the power of the PointNet model for point feature learning. To enable multiscale point feature learning, \cite{qi2017pointnet++} further designed a set abstract module and a feature propagation module for point feature downsampling and upsampling, respectively, and developed a hierarchical neural network that applies unit PointNet recursively on a nested partitioning of the input point set.

Following the great success of PointNet \citep{qi2017pointnet} and PointNet++ \citep{qi2017pointnet++}, studies have shifted their focus towards PointNet-like architectures for airborne LiDAR point cloud classification. For example, \citeauthor{yousefhussien2018multi} designed a 1D-fully convolutional network to achieve point cloud classification by inputting not only the raw coordinates but also three additional corresponding spectral features extracted from 2D georeferenced images for each point.
% Similarly, \citeauthor{wang2018deep} firstly acquired a 54-dimension feature vector by calculating an 18-dimension feature vector,  via its k-nearest neighbors, wherein k is set as 3 different values.
% Similarly, \citeauthor {wang2018deep} firstly calculated a 54-dimensional feature vector by setting three different numbers of nearest neighbor points, where for a specific number of nearest neighbor points value each point contains an 18-dimensional feature vector, which comprises of eigenvalue feature with 6 dimensions and the spin image feature with 12 dimensions.
% Similarly, \citeauthor {wang2018deep} firstly calculated a 54-dimensional feature vector by computing an 18-dimension feature vector for each point, via setting 3 different values for the number of nearest neighbors.
Similarly, \citeauthor {wang2018deep} first extracted a 54-dimensional feature vector for each point in the input point set. Those features comprise a 6-dimensional eigenvalue feature vector and a 12-dimensional spin image feature vector using 3 different settings for the parameter K of a nearest-neighbor search. Then, they developed a deep neural network with spatial pooling to aggregate the above point-based features into cluster-based features, which were further fed into a multi-layer perceptron network for point cloud classification \citep{wang2018deep}. Although these works embody methods that learn 3D point clouds directly, they still require integrated handcrafted features to improve the classification accuracy on airborne LiDAR data, which makes the task tedious and complicated.

In this paper, we proposed a directionally constrained fully convolutional neural network (D-FCN) that can be directly applied to unstructured 3D point clouds and predict the end-to-end semantic labels for arbitrary input point sizes. Furthermore, compared to other existing point cloud-based deep learning methods, the proposed model can extract representative orientation-aware features. Further, it achieves a new state-of-the-art classification performance level on airborne LiDAR point clouds while taking as input only the raw 3D coordinates and intensity; it does not require any additional handcrafted features.

\section{Methods}\label{Methods}

Our D-FCN model follows an encode-decode framework similar to the general semantic segmentation network \citep{badrinarayanan2017segnet} for 2D image segmentation (illustrated in Figure \ref{fig_arch}).

In the downsampling stage, we recursively apply our proposed directionally constrained point convolution module (D-Conv) prior to each downsampling operation for hierarchical feature embedding. In the upsampling stage, our model achieves dense feature prediction by interleaving feature propagation with our proposed D-Conv module. The core component of our segmentation network is the D-Conv module, which enables orientation aware point feature extraction. We discuss the proposed directionally constrained convolution module in Section \ref{sc_dconv} and then introduce the overall architecture of our model in Section \ref{sc_net_arch}.

\subsection{Directionally constrained point convolution}\label{sc_dconv}

Given an input point cloud $P \in \mathbf{R}^{N \times 3}$ and a feature matrix $F\in \mathcal{R}^{N \times d_{in}}$ of one hidden layer, where $N$ represents the number of points in the input point set and $d_{in}$ denotes the dimension of the input feature map, our D-Conv module generates the output $O\in \mathcal{R}^{N \times d_{out}}$, and assigns a new $d_{out}$-dimension feature vector to each point.
Here, similar to the conventional convolution operator, the proposed D-Conv module covers a small receptive field and is responsible for local feature extraction. We design our directionally constrained point convolution module following the construction patterns of the conventional convolution operator. Two steps are involved in the convolution module: we construct a local receptive field foreach point and then apply the convolution operation to the feature values within the receptive field using the kernel weights.

\begin{figure}[t]
\centering
\includegraphics[width=8cm]{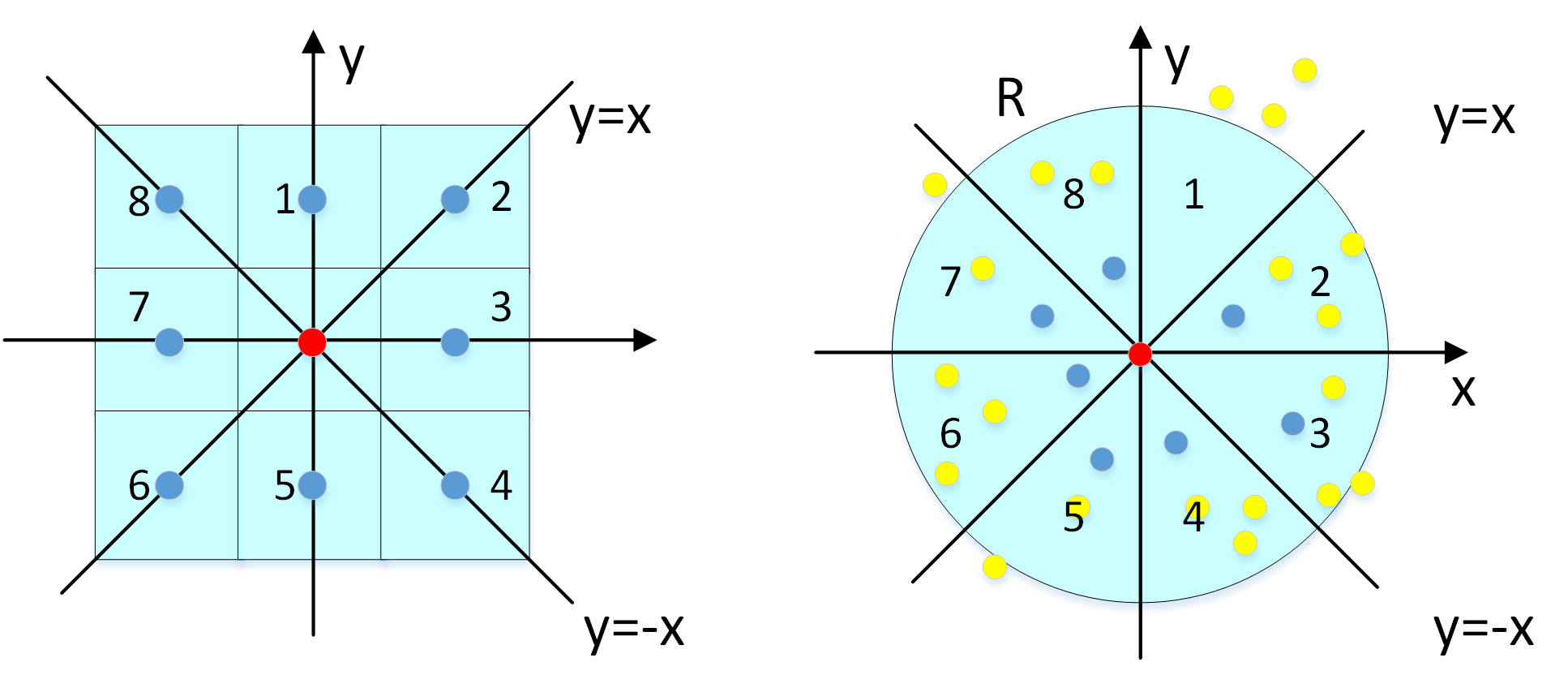}
\caption{Conventional convolution (left) with a kernel size of 3x3 and the proposed directionally constrained K-nearest neighbor operation (right). For both convolution operations, the 2D space is evenly divided into 8 triangle areas each of which covers an angle of 45 degrees. The input point is at the origin, marked in red, while the blue and yellow points indicate the selected supporting points and the nonsupporting points, respectively.}
\label{fig_dknn}
\end{figure}

\subsubsection{Directionally Constrained Nearest Neighbor}~\\
The conventional convolution operator starts by constructing a local receptive field to enable local feature extraction. Figure \ref{fig_dknn} (left) gives an example of conventional 3x3 convolution; given a central point (marked in red), the receptive field covers an area of 3x3 gridded pixels, and each pixel (marked in blue) can be regarded as the nearest neighbor in each of the 8 evenly divided directions, except for the central point itself.

Similarly, the proposed D-Conv module formulates its receptive field by finding the nearest neighbor points in $N_d$ (by default, we set $N_d$ to 8) evenly divided directions. More importantly, instead of dividing the original 3D point space into cone subspaces, we first project all the input 3D points onto an xy plane and then formulate the receptive field in 2D point space.

\begin{figure}[t]
\centering
\includegraphics[width=8cm]{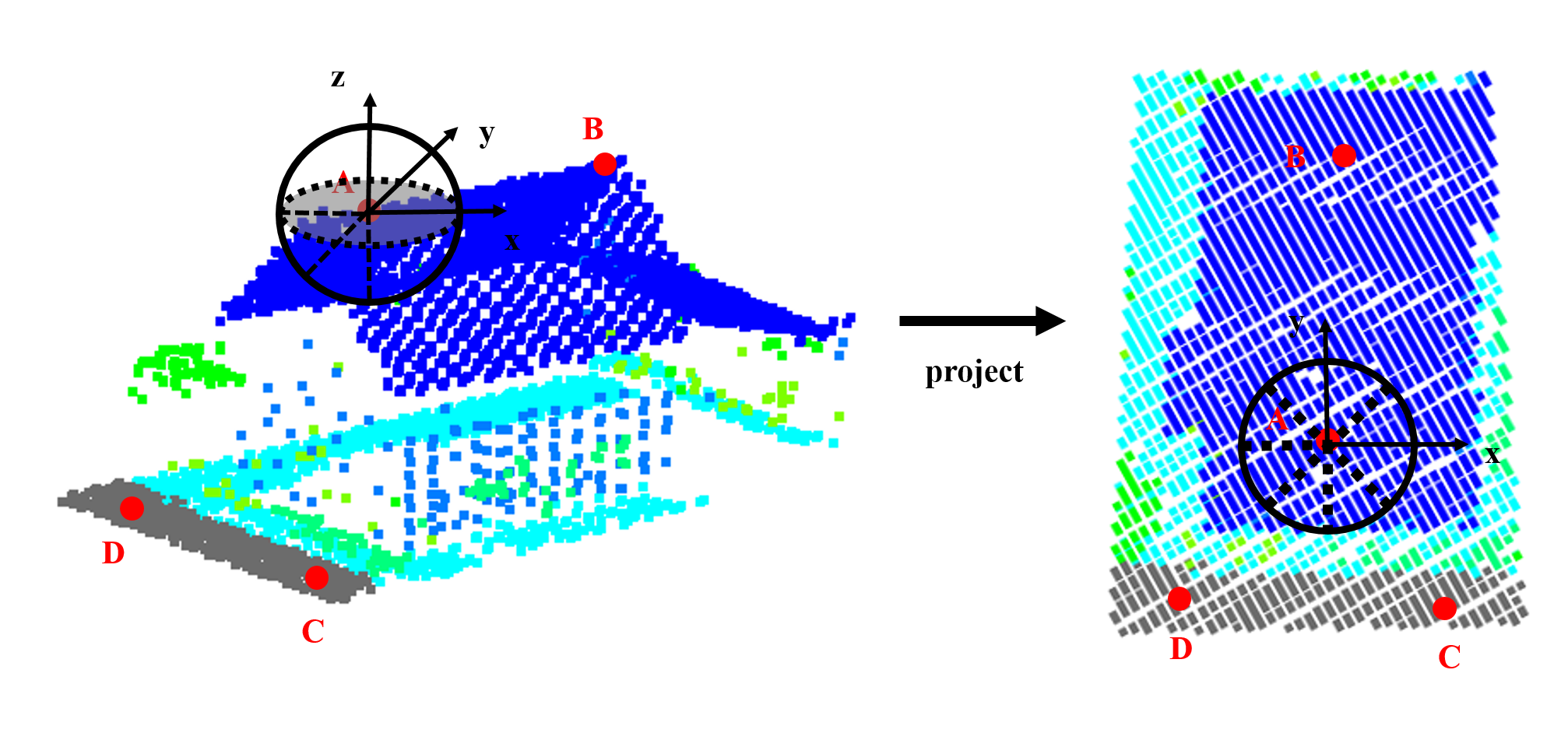}
\caption{3D neighborhood search (left) and projected 2D neighborhood search (right).}
\label{fig_prj}
\end{figure}

The reason why we do not divide the 3D space based on the original xyz coordinates is that the 3D point cloud captured by airborne LiDAR has a larger variance in the horizontal direction (xy coordinates) than in the vertical direction (z coordinates). Therefore, projecting points to the xy plane during the neighborhood search enables more representative neighbor points in the xy direction while paying less attention to points projected to nearby locations on the ground plane that share the same semantic labels. This is the reason why previous works such as \citep{zhao2018classifying} and \citep{yang2018segmentation} directly project the original 3D point cloud into 2D images and use conventional 2D CNN to predict the semantic label for each point. Figure \ref{fig_prj} shows an illustration of this statement. Building rooftop points such as $A, B$ have no upper neighbor points, while the ground points such as $C, D$ have no lower neighbor points. Therefore, searching directionally constrained nearest neighbors in the original 3D space causes many vacant subspaces and introduces fewer informative neighborhood points. In contrast, the proposed 2D search strategy guarantees fewer vacant subspaces and enables richer neighborhood information. Another advantageous property of this 2D search strategy is that dividing a 2D space into triangular areas is much easier and more computationally efficient than dividing a 3D space into cones. We verify the advantages of our 2D search strategy over the 3D search strategy in Section \ref{Experiment_Dconv}.

After projecting the neighborhood space onto the xy plane, we divide the 2D plane originating at each central point $p_i$ into $N_d$ triangle areas. Without special declarations, we set $N_d$ to 8 in this paper. Then, we search $K$-nearest neighbor points in each area within a radius of $R$ to formulate the receptive field of the central point (see Figure \ref{fig_dknn} (right) for illustration). In general, a larger $K$ enables a more reliable and stable representation of the captured features; however, a larger $K$ can also lead to overfitting. The effects of different values of $N_d$ and $K$ are discussed in Section \ref{Experiment_Model_Parameters}. Moreover, the distant points tend to provide little information for describing the local patterns at the central point $p_i$. When no point exists within the search radius $R$ in a triangle area (e.g., area `1' in the right-hand figure), we duplicate the central point $p_i$ as the nearest neighbor of itself. After searching for each central point, the proposed directionally constrained nearest neighbor generates $N_d$ groups of $K$ points in a fixed order.

Although we project all the 3D points onto the xy plane, the z coordinate feature is maintained for point feature encoding, as discussed in the next section.

\begin{figure}[t]
\centering
\includegraphics[width=8cm]{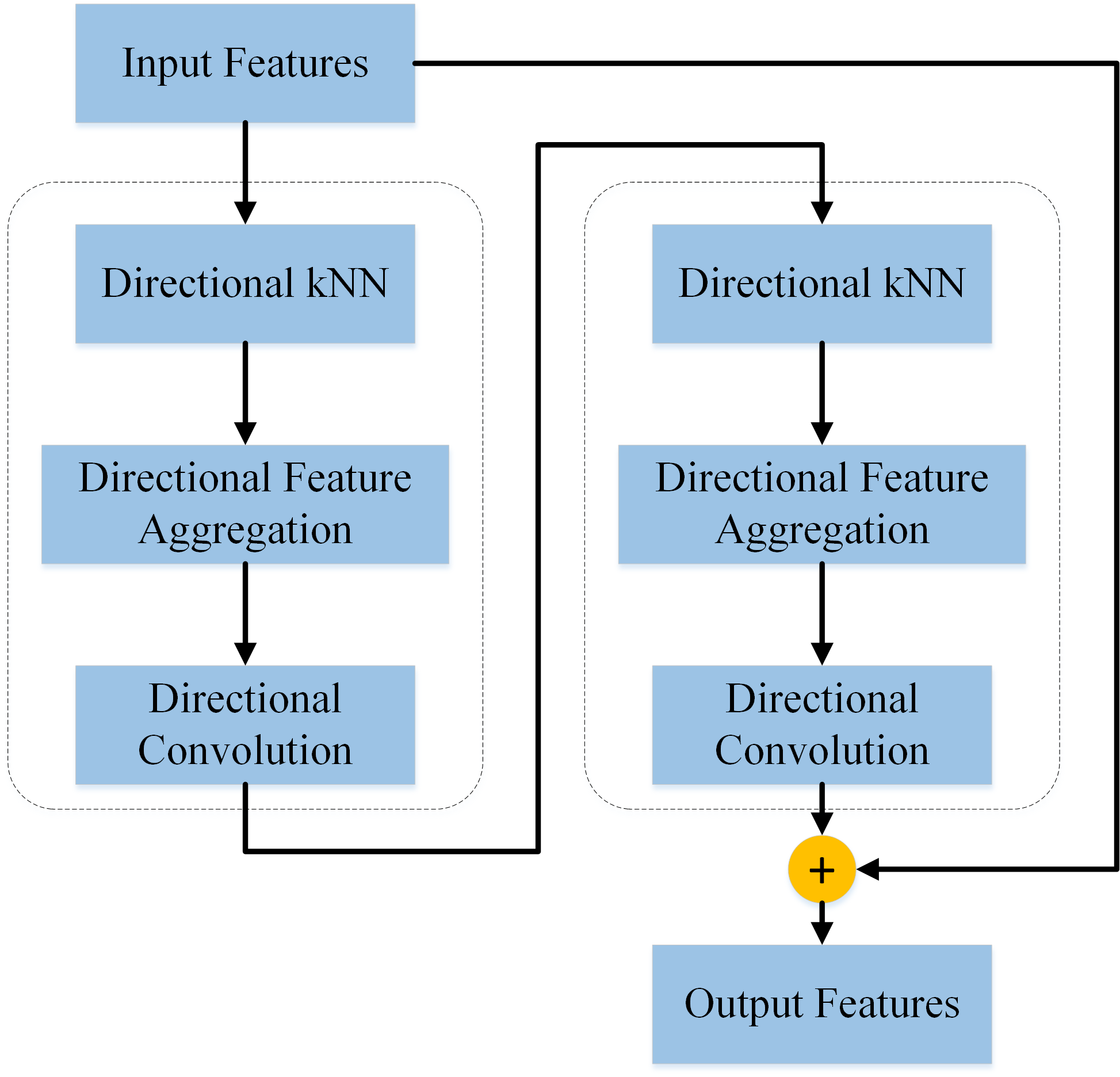}
\caption{Illustration of the proposed D-Conv module. The dashed box indicates one convolution block. In this paper, we use two convolution blocks in each D-Conv module. A `+' indicates an element-wise summation operation.}
\label{fig_dconv}
\end{figure}

\subsubsection{Orientation-Aware Point Feature Encoding}~\\
In a conventional convolutional layer, the convolution operation for each kernel preserves the same order when sliding through the whole input feature maps, thus ensuring the translation invariance property. An important prerequisite for this operation is that in regular-gridded feature inputs, a well-defined order exists for organizing the neighbor points within the receptive field. Figure \ref{fig_dknn} (left) gives an example of the ordering of neighbor points starting from 1 to $N_d$.

However, 3D point clouds are presented in an unordered and unstructured format. No regular order exists for organizing the neighbor points. Early works partitioned the input 3D point space into gridded voxels to enable an order formulation. Recent PointNet-like works commonly apply unordered operations (e.g., max pooling \citep{qi2017pointnet++,wang2018dynamic,shen2018mining,jiang2018pointsift}) to aggregate the features of all the neighbor points when formulating the local feature for each central point. However, these unordered operations ignore order information that can capture the spatial structure of the input points; consequently, they also cause information loss (e.g., the max pooling operation discards all inputs except for the maximum).

To enable orientation-aware point feature learning, our D-Conv module is therefore expected to maintain a well-defined order when applied to each point, which can be easily achieved by adopting the order of each triangle area when formulating the receptive field (see Figure \ref{fig_dknn} (right) for illustration). In each of the $N_d$ triangle areas, the features of the $K$ neighbor points are first aggregated to generate the directionally aware features, resulting in a $1\times N_d$ feature vector for each central point. Given central point $p_i$ and $N_d$ groups of $K$ neighbor points, the directionally aware feature vector $\tilde{f} \in \mathcal{R}^{d_{out}}$ of the $j$th triangle area is calculated as follows:
\begin{equation}
\tilde{f}_{ijc} = \sum_{k=1}^{K} w_{ck}*f_{ijk}, \quad c \in \{1,2,...,d_{out}\}, j \in \{1,2,...,N_d\},
\end{equation}
where $k\in \{1,2,...,K\}$ denotes the neighbor index in each triangle area, and $w_{ck}$ denotes the corresponding weight parameter of the $c$th output channel. This aggregation function can easily be implemented using a $1\times K$ convolution with a stride of $K$ and $d_{out}$ output channels on the grouped input features of dimension $B\times N \times (N_d \times K) \times d_{in}$, where $B$ denotes the batch size. The output tensor dimension is $B \times N \times N_d \times d_{out}$.

Then, our orientation-aware point convolution is formulated as a $1\times N_d$ convolution with a stride of $N_d$ and $d_{out}$ output channels. Given central point $p_i$ and the directionally aware feature vectors $\tilde{f}$, the output feature $O_i \in \mathbf{R}^{d_{out}}$ at central point $p_i$ is calculated as
\begin{equation}
O_{ih} = \sum_{c=1}^{d_{out}} \sum_{j=1}^{N_d} w_{hj}*\tilde{f}_{ijc}, \quad h \in \{1,2,...,d_{out}\},
\end{equation}
where $w_{hj}$ indicates the weight parameter for the $j$th triangle area of the $h$th output channel. The output tensor dimension is $B \times N \times 1 \times d_{out}$.

In sum, our D-Conv module starts by finding the $K$-nearest neighbors in each of the $N_d$ triangle areas partitioned on the xy plane. Then, we aggregate the features in each triangle area by using a $1 \times K$ convolution with a stride of $K$. Next, orientation-aware point convolution is applied to extract the point descriptors for each central point using a $1 \times N_d$ convolution with a stride of $N_d$. Finally, we add the output features to the input using an element-wise summation operation. The overall architecture of our D-Conv module is illustrated in Figure \ref{fig_dconv}. In Figure \ref{fig_dconv}, we repeat the convolution block twice to enable more representative feature learning.

Note that our D-FCN module does not change the size of the input point set. In the next section, we introduce downsampling and upsampling blocks to enable multiscale point feature aggregation and propagation.

\begin{figure*}[t]
\centering
\includegraphics[width=17cm]{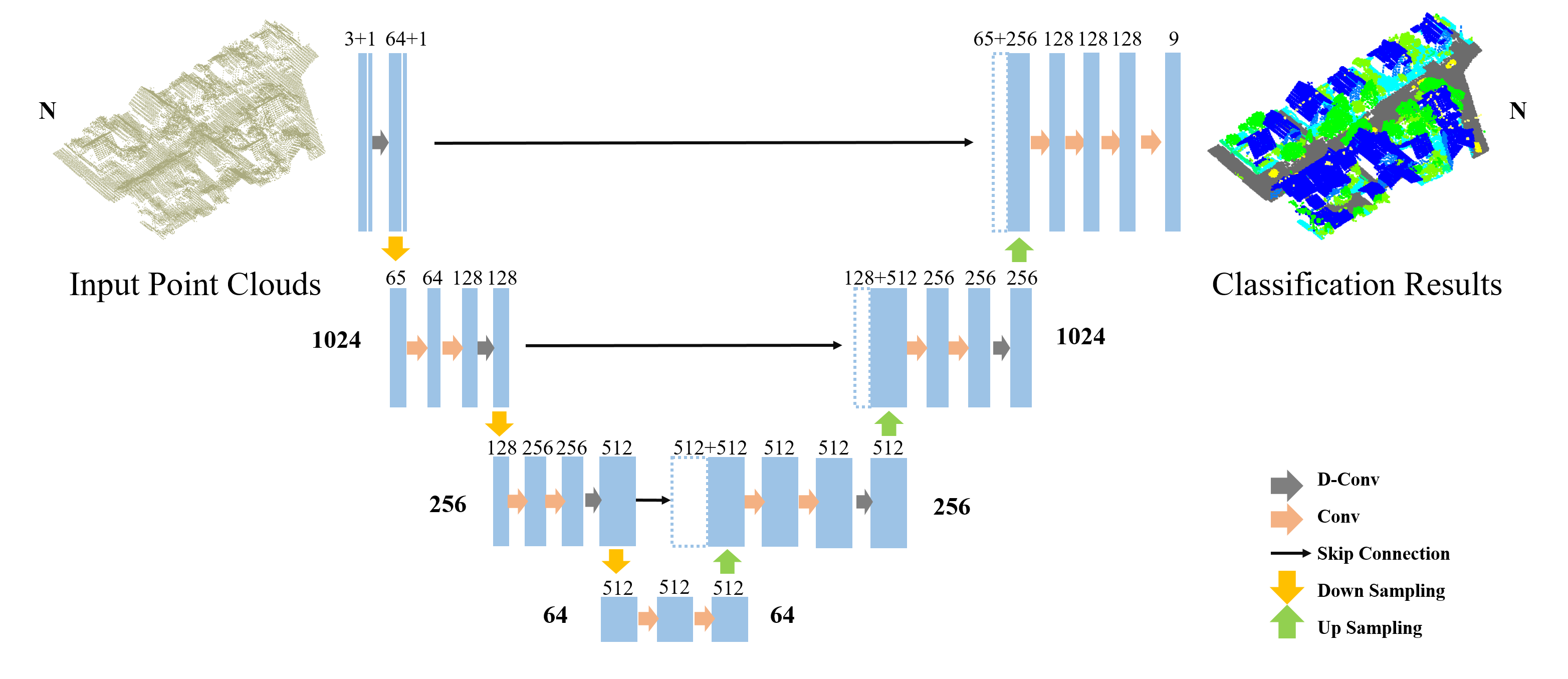}
\caption{
Illustration of our D-FCN network architecture. The network includes both a downsampling path and an upsampling path. Our D-Conv module (marked in gray) is inserted before each downsampling operation (marked in yellow) and each upsampling operation (marked in green). Skip connections are used to propagate information from the downsampling path to the upsampling path.
}
\label{fig_arch}
\end{figure*}

\subsection{Network architecture}\label{sc_net_arch}
\subsubsection{Downsampling and Upsampling blocks}~\\
To enable multiscale high-level feature learning, in this study, we adopt downsampling modules to extract multiscale features and upsampling modules to obtain point features for all the original points. These modules were introduced in PointNet++ \citep{qi2017pointnet++} to enable hierarchical point feature learning. Here, we provide a brief introduction to these two modules.

A downsampling module takes a $N_l \times d_{l}$ feature matrix as input, where $N_l$ represents the number of input points and $d_l$ indicates the feature dimension, and outputs a $N_{l+1} \times d_{l+1}$ feature matrix with a reduced number of points $N_{l+1}$ ($N_{l+1} \le N_{l}$) and a feature dimension of $d_{l+1}$. The downsampling process is implemented by farthest point sampling to select a subset of points such that each selected point is the most distant point from the set of all the other points. After point downsampling, the embeddings for each selected point are calculated by passing the features of its neighbor points through a convolution module. A ball query or K-nearest neighbor search strategy can be used to define these neighbors, and the max-pooling function is used to aggregate the features in the convolution module. By default, we set the neighborhood size to 32 in this study.

An upsampling module propagates the point features from $N_{l}$ points to $N_{l-1}$ points (with $N_{l-1} \ge N_l$) using distance-based interpolation. Specifically, the features of the points dropped during the downsampling process are calculated using the inverse distance weighted average based on the k-nearest neighbors. As the default setting, for each point to be interpolated, we search for 32 neighbor points in the input points to produce the point feature for these dropped points. The interpolated features from the $N_{l-1}$ points are then concatenated with the features obtained during the downsampling process with the same number of points.

\subsubsection{Overall architecture}~\\
Our point segmentation network is organized in a U-Net-like structure, as shown in Figure \ref{fig_arch}. It takes the 3D coordinates and the intensity feature as input and outputs a semantic label for each point. During downsampling, three successive downsampling operations are applied to shrink the size of the point set to 1,024, 256, 64. Then, during the upsampling stage, three successive upsampling operations are utilized to generate dense feature prediction. The point set size is increased to 256, 1,024, 8,192 points in each upsampling stage. We insert our D-Conv module is inserted before each downsampling operation and each upsampling operation. As a default setting, the search radius $R$ for each sampling level was set to 2, 5, 10 from bottom to top. Finally, the point features of the last upsampling layer are input into a fully connected layer to produce a semantic label for all the input points. Moreover, to incorporate the low-level information in the downsampling stage, we add skip connections between the downsampling and upsampling stages. The low level-level features from the downsampling stage are concatenated with the feature matrix of the same point set size in the upsampling stage. Previous works in 2D segmentation tasks have frequently reported that employing skip connections boost the convergence speed and improve performance \citep{ronneberger2015u,badrinarayanan2017segnet}.

\subsection{Evaluation metrics}\label{Methods_evaluation}

Following the standard convention of the ISPRS contest, precision, recall, F1 score, and overall accuracy (OA) are commonly used to evaluate the performance of 3D point cloud labeling. In general, overall accuracy is used to assess the overall classification accuracy for all categories, which is defined as the percentage of correctly classified points in the total test points. Moreover, the F1 score takes the precision and recall of the classification model into account and is generally more suitable for cases where the categories are unevenly distributed. The precision, recall and F1 score for each category are defined as follows:

\begin{equation}
precision = \frac{TP}{TP+FP}
\end{equation}

\begin{equation}
recall = \frac{TP}{TP+FN}
\end{equation}

\begin{equation}
F1 = 2*\frac{precision*recall}{precision+recall},
\end{equation}

where TP (true positive), FP (false positive), and FN (false negative) respectively indicate the positive tuples that were correctly labeled by the classifier, the negative tuples that were incorrectly labeled as positive, and the positive tuples that were mislabeled as negative.

\section{Experiments and Discussion}\label{Experiments}

In this section, we present the extensive experiments conducted to evaluate the performance of the proposed model for Airborne LiDAR point cloud classification. In Section \ref{Experiment_Data}, we introduce the experimental dataset. In Section \ref{sc_preprocessing}, we present two processing strategies, patch sampling and class balance. The experimental setup is given in \ref{Experiment_Setup}. In Section \ref{Experiment_Dconv}, we evaluate the effect of the proposed directionally constrained point convolution module. In Section \ref{Experiment_Model_Parameters}, we discuss the effect of different hyper-parameter configurations. The classification results of our model with optimal hyperparameters are given in Section \ref{Experiment_Result}. Finally, in Section \ref{Experiment_Comparison }, we compare our model with other state-of-the-art models for airborne LiDAR point cloud classification.

\subsection{Experimental dataset}\label{Experiment_Data}

To evaluate the performance of the proposed method, we conducted experiments on the International Society for Photogrammetry and Remote Sensing (ISPRS) 3D labeling dataset \citep{niemeyer2014contextual}. This dataset includes three test areas that include reference data for various object classes. The airborne LiDAR point clouds were captured by a Leica ALS50 system, at an average flying height of 500 meters above the ground and a field of view of 45 degrees \citep{cramer2010dgpf}. The LiDAR points were labeled with 9 semantic categories, including powerline, low vegetation (low\_veg), impervious surface (imp\_surf), car, fence/hedge (fen/hed), roof, facade, shrub, and tree.
The spatial 3D coordinates (XYZ), intensity values, number of returns and point labels were stored in LiDAR point files.

\begin{figure*}
\centering
\includegraphics[width=17cm]{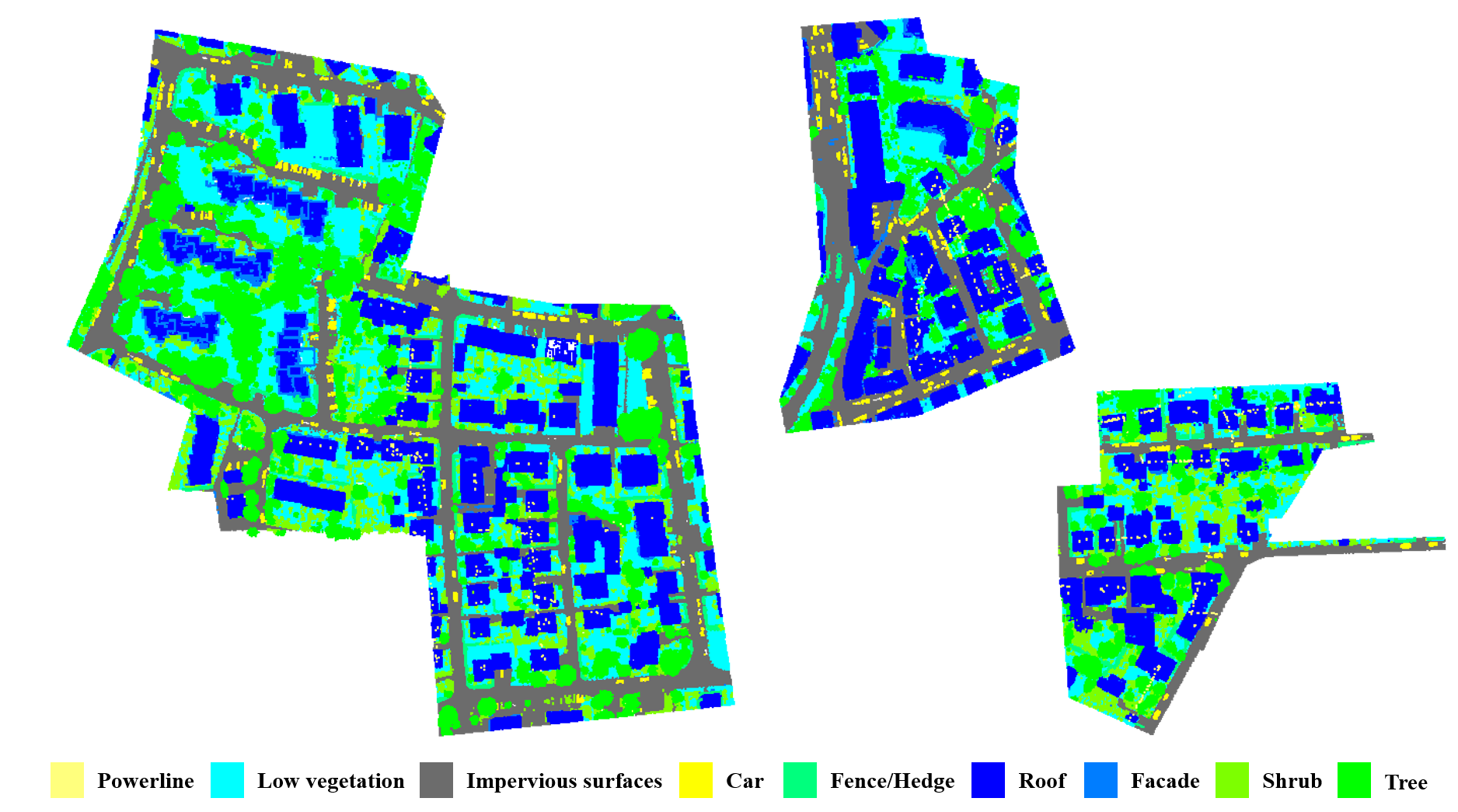}
\caption{ISPRS 3D labeling dataset. From left to right, Scene I (left) is used for model training and Scenes II and III (right) are used for model evaluation. The point categories in each scene are rendered in the colors provided by the contest, see the legend at the bottom. Best viewed in color.
}
\label{fig_Dataset}
\end{figure*}

Following the standard setting of the ISPRS 3D labeling contest, the dataset was divided into two parts according to the scene areas. The first scene (see Figure \ref{fig_Dataset} left) with 753,876 points was used as the training dataset, and the other two scenes (see Figure \ref{fig_Dataset} right) with 411,722 points as the test dataset. The detailed category distribution of each scene is shown in Table \ref{table_Dataset_number}. Note that the test dataset labels were used only for model evaluation, which is unknown during the model training stage.

\begin{table*}[h]
\begin{center}
\caption{Point category distribution of each scene.}
\label{table_Dataset_number}
\begin{tabular}{l c c c c}

\hline
Categories & Scene I & Scene II & Scene III & Scene II\&III Total
\\
\hline
Powerline & 546 & 218 & 382 & 600\\
Low vegetation & 180,850 & 38,104 & 60,586 & 98,690\\
Impervious surfaces & 193,723 & 53,255 & 48,731 & 101,986\\
Car &  4,614 & 2,184 & 1,524 & 3,708\\
Fence/Hedge & 12,070 & 1,479 & 5,943 & 7,422\\
Roof & 152,045 & 53,445 &  55,603 & 109,048\\
Facade & 27,250 & 5,983 &  5,241 & 11,224\\
Shrub & 47,605 & 1,414 & 23,404 & 24,818\\
Tree & 135,173 & 23,915 & 30,311 & 54,226\\
\hline
Total & 753,876 & 179,997 & 231,725 & 411,722\\       
% \hline

\hline

\end{tabular}
\end{center}
\end{table*}

\subsection{Class balance and data augmentation strategies}\label{sc_preprocessing}

As shown in Table \ref{table_Dataset_number}, the category distribution of each scene was extremely uneven, as indicated by the number of points in each category. For example, in scene I, merely 546 points are labeled as powerline, while 193,723 points are labeled as impervious surfaces. Conducting training directly on this unbalanced dataset can potentially have a severe negative impact on the overall performance of deep networks, as discussed in a recent technique report \citep{masko2015impact}. To address this issue, we added a category-specific weight coefficient for each category to the loss function of our D-FCN model. This strategy was designed to encourage our model to pay more attention to those points in the minority categories. The balance weight for each category was calculated by the logarithm function of the percentage of each category and then performing the reciprocal operation, formulated as Equation \ref{layer_function}.

\begin{equation}
{W}_{c} = \frac{1}{\ln (\alpha + \frac{{N}_{c}}{{\sum_{c=1}^{C} {N}_{c}}})},
\label{layer_function}
\end{equation}

where $W_c$ refers to the weight of the $c$th category, $N_c$ represents the number of points of the $c$th category, $C$ denotes the total number of categories, and $\alpha$ denotes the coefficient for class balance. The value of $\alpha$ is discussed in Section \ref{Experiment_Model_Parameters}. After integrating the class balance weights, our final loss function is defined as follows:
\begin{equation}
\mathcal{L} = \sum_{i=1}^{N} w_i * \sum_{c=1}^{C} [y_{ic} logp_{ic} + (1-y_{ic}) log(1-p_{ic}) ]
\label{loss_function},
\end{equation}
where $N$ denotes the number of sampling points in each training block, $y_{ic}$ and $p_{ic}$ denote the ground truth label and predicted probability of the $i$th point for the $c$th category, and $w_i$ denotes the balance weight for the $i$th sampling point, calculated as $w_i:= W_{c=y_i}$.

Moreover, to improve the generalizability, we introduce a data augmentation technique to further enhance the robustness of our proposed model. Considering that our training scene has an irregular boundary, we referenced the data augmentation algorithm introduced in PointNet++ \citep{qi2017pointnet++}. During training, we randomly select a 30 m*30 m*40 m cuboid region from the whole scene and then randomly choose 8,192 points from the cuboid as the model input. To further reduce the risk of overfitting and make the model more robust, the selected 8,192 points are randomly dropped during the training stage. By default, the dropout ratio is set to 12.5\% in the following sections. With respect to the testing dataset, the scenes were segmented into blocks of 30 m*30 m grids in the horizontal direction (see Figure \ref{fig_Testset_split}). Note that small blocks generated at the edge of the scene were merged into the surrounding large blocks to ensure the integrity of each block. Also note that the test blocks can have substantially different numbers of points. Fortunately, due to the fully convolution nature of the proposed D-FCN model, all the points in each test block can be input directly into the model for point classification.

\begin{figure}[h]
\centering
\includegraphics[width=8cm]{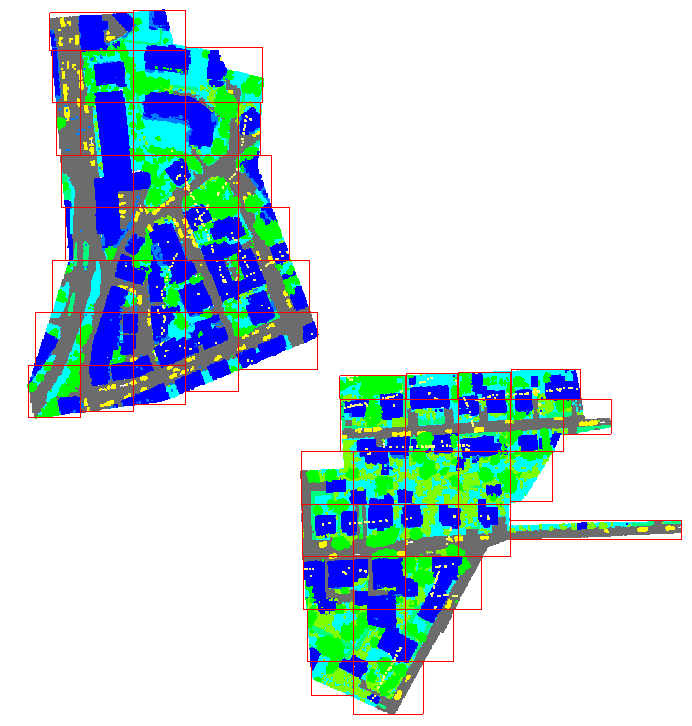}
\caption{The blocks of the test dataset (i.e. Scene II Scene III) were segmented into 30m * 30m grids in the horizontal direction and small blocks broken at the edges were merged into the surrounding large blocks.
}
\label{fig_Testset_split}
\end{figure}

\subsection{Experimental setup}\label{Experiment_Setup}

We implemented our D-FCN method using the TensorFlow framework. During network training, we set the batch size to 6 and used the Adam optimizer with an initial learning rate of 0.01. We divided the learning rate by 2 every 3,000 steps. Training our model for 1,000 epochs until convergence required approximately 10 hours running on a TESLA K80 GPU.

During the test stage, we fed each test block directly into our model and generated the classification results in one network forward pass. We merged the prediction results of each block to obtain the final point cloud segmentation results, see Figure \ref{fig_Result_overview}. Our code will be released at \url{https://github.com/lixiang-ucas/D-FCN}.

\subsection{Effect of directionally constrained point convolutions}\label{Experiment_Dconv}

To explore the effect of the directionally constrained point convolutions module and further determine the optimal directional searching strategy of the neighbor points, we designed four models ( i.e., a model with no directionally constrained point convolution module, a model with an eight-directional search in 3D space, a model with a four-directional search in 2D space and a model with an eight-directional search in 2D space) and conduct point cloud classification using identical network configurations.

\begin{figure}[h]
\centering
\includegraphics[width=8cm]{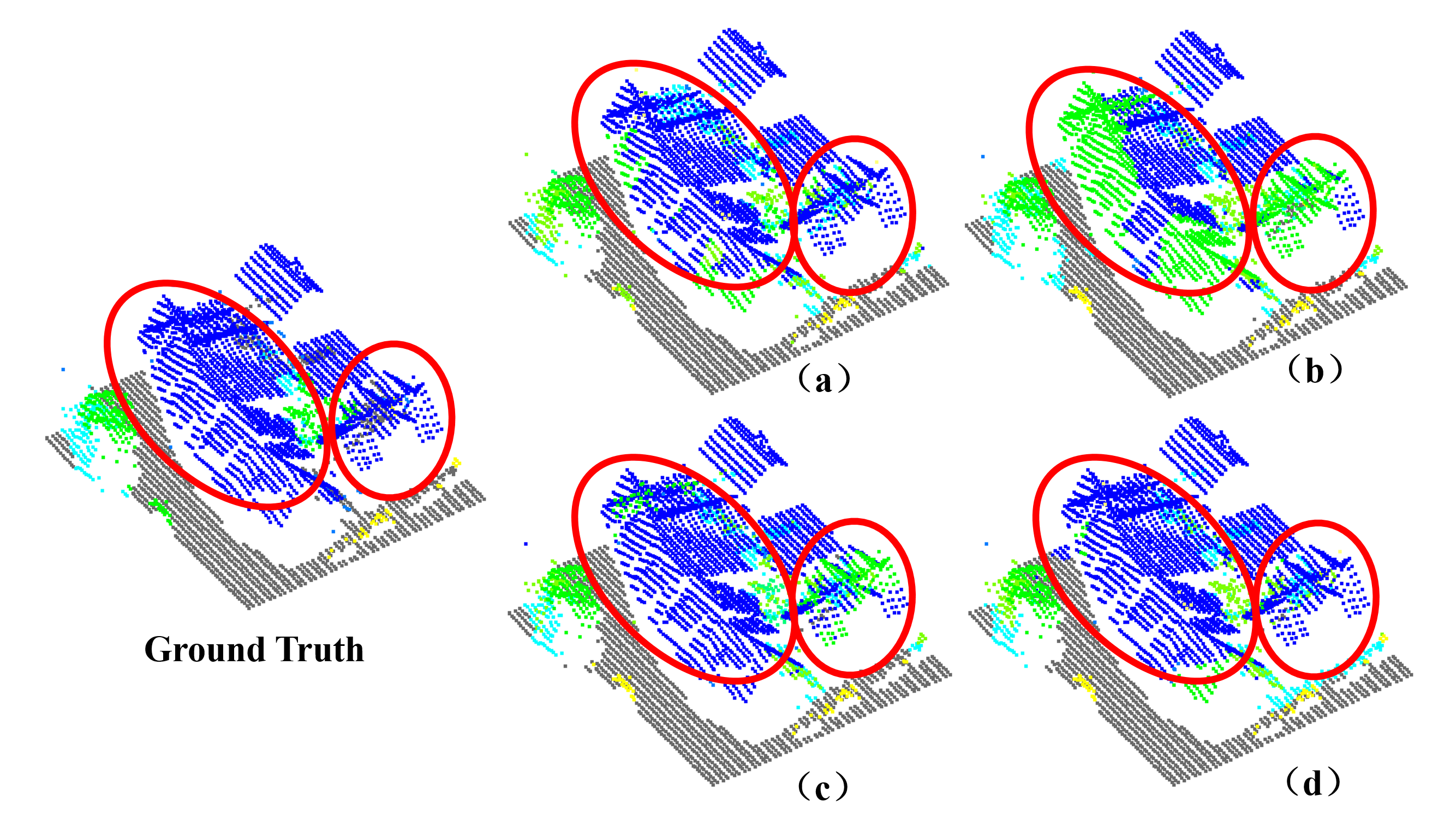}
\caption{The classification results under different configurations of the directionally constrained point convolution module. Model with no directionally constrained point convolutions (a); model with eight directions in 3D space (b); model with four directions in 2D space (c); and model with eight directions in 2D space (d). The circled parts highlight the most obvious differences among the classification results of the four models.
}
\label{fig_effect_dconv}
\end{figure}

Here, as a default configuration, the number of sampling points $N$ was set to 2,048, and we sampled 1 point in each directionally constrained area (i.e., $K=1$). The $\alpha$ class balance coefficient was set to 1.2.
The classification results of all four compared models are listed in Table \ref{table_directions}. Here, we report both the overall accuracy and the average F1 score. The corresponding qualitative results are shown in Figure \ref{fig_effect_dconv}.

In Figure \ref{fig_effect_dconv}, the model without a directionally constrained point convolution module misclassified some roof points into the tree category. More importantly, the proposed models with directionally constrained searching in 2D space outperformed the model with directionally constrained search in 3D space regardless of the number of directions used, as indicated by the overall accuracy and average F1 score. This is probably because the projected 2D searching strategy enables richer informative neighbor points than does the 3D searching strategy, as discussed in Section \ref{sc_dconv}. Moreover, comparing the results of the last two rows in Table \ref{table_directions}, it is clear that instead of dividing the 2D space into 4 directionally constrained areas, using 8 directionally constrained areas leads to more representative neighbor points and achieves better classification results. Hereafter, our directionally constrained point convolution module adopts an 8-directional search in 2D space.

\begin{table}[h]
\begin{center}
\caption{The overall accuracy (OA) and average F1 scores of the classification results of the models under different neighborhood partition strategies. From top to bottom are the model with no directionally constrained point convolution module and the models employing eight-directional search in 3D space, four-directional search in 2D space and eight-directional search in 2D space. The boldface entries indicate the model with the best performance. }
\label{table_directions}
\begin{tabular}{l c c}
\hline
Neighborhood partition strategy & OA & Average F1
\\
\hline
no partition & 0.817 & 0.663\\
3D space(8 directions) & 0.818 & 0.668\\
2D space(4 directions) & 0.820 & 0.669\\
\textbf{2D space(8 directions)} & \textbf{0.833} & \textbf{0.699 }\\
\hline
\end{tabular}
\end{center}
\end{table}

\subsection{Model hyperparameters}\label{Experiment_Model_Parameters}
In this section, we conduct extensive experiments to determine the optimal model hyperparameters, including the number of sampling points $N$ in each training block, the number of neighbor points $K$ in each triangle area and the class balance coefficient $\alpha$. Average F1 score was adopted as the evaluation metric to select the best hyperparameters.

\subsubsection{Number of sampling points and neighbor points}~\\
As mentioned in Section \ref{sc_dconv}, the number of neighbor points $K$ plays an important role in model performance. Selecting an appropriate number of these points enables a more reliable and stable representation of the captured features. Meanwhile, the number of sampling points $N$ in each training block controls the density of the input point set. In a sparse point set (i.e., where $N$ is small), each central point tends to have fewer candidate points that can be selected within the searching radius $R$. Therefore, we discuss the number of sampling points $N$ and neighbor points $K$ simultaneously to determine their optimal configurations.

In our experiments, we randomly sampled 2,048, 4,096, and 8,192 points from each training block as the input to our D-FCN model and searched 1, 2, and 4 neighbor points from each of the 8 directionally constrained triangle areas in our D-Conv module. In these experiments, the class balance coefficient $\alpha$ was 1.2. The classification performances were evaluated regarding both overall accuracy (OA) and average F1 score, as listed in Table \ref{table_sample_neighbor}.

As shown in Table \ref{table_sample_neighbor}, the model achieved a better classification performance by taking 1 nearest neighbor point in each direction (i.e., $K=1$) when fewer points were sampled in each training block (e.g., $N=2,048$). This result probably occurs because a larger neighborhood size involves more redundant information when the point set is sparse. As the number of sampling points increases (e.g., $N=8,192$), the increase in neighbor points ($K=2$) helps to obtain more stable features that improve the classification results. Note that using too many neighbor points ($K=4$) can have a negative effect on the classification performance when $N$ is sufficiently large. Our model achieved its best performance when $N$ and $K$ were set to 8,192 and 2, respectively, with an overall accuracy of 82.2\% and an average F1 score of 70.7\%. Hereafter, we use this configuration for $N$ and $K$ in our experiments.

\begin{table}[h]
\begin{center}
\caption{The overall accuracy (OA) and average F1 score of model classification results under different numbers of sampling points ($N$) and neighbor points ($K$). The boldface text indicates the model with the best performance. }
\label{table_sample_neighbor}
\begin{tabular}{l c c c}
\hline
$N$ & $K$ & OA & Average F1
\\
\hline

2,048 & 1 & \textbf{0.833}  & 0.699 \\
2,048 & 2 & 0.819  & 0.687 \\
2,048 & 4 & 0.819  & 0.679 \\
4,096 & 1 & 0.827  & 0.698 \\
4,096 & 2 & 0.828  & 0.698 \\
4,096 & 4 & 0.831  & 0.700 \\
8,192 & 1 & 0.828  & 0.704 \\
\textbf{8,192} & \textbf{2} & 0.822  & \textbf{0.707} \\
8,192 & 4 & 0.827  & 0.701 \\

\hline
\end{tabular}
\end{center}
\end{table}

\subsubsection{Weight coefficient of data balance}~\\
In this section, we conducted comparative experiments with different class balance coefficient ($\alpha$) configurations; namely, a model with no data balancing, and models with $\alpha$ set to 1.1, 1.2, 1.3, 1.4 and 1.5. Table \ref{table_balance} shows the classification results under different $\alpha$ configurations. 

As shown in Table \ref{table_balance}, our model achieves the best overall accuracy (84.6\%) with no class balancing strategy. However, its average F1 score is lower than the models with class balance. This result is probably caused by the uneven distribution of the input class categories (see Table \ref{table_Dataset_number}). When using no class balancing strategy, our model paid more attention to the categories with more points (e.g., low vegetation, impervious surfaces) and thereby achieved higher accuracy for these majority categories. However, the categories with fewer points (e.g., powerline, car) may be ignored during the model training, resulting in lower accuracy for these categories.

When the class balancing strategy was used, the models improved the average F1 score. The highest average F1 score of 70.7\% was achieved by the model where the class balance coefficient $\alpha$ was set to 1.2.

\begin{table}[h]
\begin{center}
\caption{Classification results with different values of the class balance coefficient $\alpha$. The overall accuracy (OA) and average F1 score are reported. The boldface text indicates the model with the best performance.}
\label{table_balance}
\begin{tabular}{l c c}
\hline
$\alpha$ & OA & Average F1
\\
\hline
NA & \textbf{0.846}  & 0.694 \\
1.1 & 0.820  & 0.703 \\
\textbf{1.2} & 0.822  & \textbf{0.707} \\
1.3 & 0.829  & 0.697 \\
1.4 & 0.828  & 0.701 \\
1.5 & 0.834  & 0.695 \\
\hline
\end{tabular}
\end{center}
\end{table}

\begin{table*}
\begin{center}
\caption{The classification confusion matrix of our directionally constrained fully convolutional neural network (D-FCN) model. Precision/correctness, recall/completeness, and F1 score are also reported. Our model obtained an overall accuracy of 82.2\% and an average F1 score of 70.7\%. }
\label{table_confusion_matrix}
\begin{tabular}{l c c c c c c c c c}

\hline
Categories & power & low\_veg & imp\_surf & car & fence\_hedge & roof & facade & shrub & tree 
\\
\hline
power & \textbf{0.690}  & 0.003  & 0.000  & 0.000  & 0.000  & 0.143  & 0.053  & 0.003  & 0.107 \\ 
low\_veg & 0.000  & \textbf{0.771}  & 0.069  & 0.001  & 0.004  & 0.011  & 0.004  & 0.116  & 0.023 \\ 
imp\_surf & 0.000  & 0.086  & \textbf{0.902}  & 0.001  & 0.000  & 0.005  & 0.001  & 0.006  & 0.001 \\ 
car & 0.000  & 0.053  & 0.024  & \textbf{0.714}  & 0.038  & 0.025  & 0.005  & 0.125  & 0.016 \\ 
fence\_hedge & 0.000  & 0.094  & 0.015  & 0.008  & \textbf{0.264}  & 0.013  & 0.005  & 0.483  & 0.119 \\ 
roof & 0.001  & 0.014  & 0.000  & 0.000  & 0.000  & \textbf{0.907}  & 0.020  & 0.018  & 0.040 \\ 
facade & 0.002  & 0.061  & 0.004  & 0.001  & 0.004  & 0.116  & \textbf{0.575}  & 0.138  & 0.099 \\ 
shrub & 0.000  & 0.104  & 0.006  & 0.005  & 0.016  & 0.037  & 0.017  & \textbf{0.626}  & 0.189 \\ 
tree & 0.000  & 0.010  & 0.000  & 0.001  & 0.004  & 0.013  & 0.011  & 0.139  & \textbf{0.822} \\ 
\hline
Precision/Correctness & 0.718  & 0.836  & 0.927  & 0.862  & 0.621  & 0.954  & 0.637  & 0.364  & 0.767 \\ 
Recall/Completeness & 0.690  & 0.771  & 0.902  & 0.714  & 0.264  & 0.907  & 0.575  & 0.626  & 0.822 \\ 
F1 score & 0.704  & 0.802  & 0.914  & 0.781  & 0.370  & 0.930  & 0.605  & 0.460  & 0.794  \\ 
\hline

\end{tabular}
\end{center}
\end{table*}

\subsection{Classification results}\label{Experiment_Result}
\begin{figure*}[p]
\centering
\includegraphics[width=17cm]{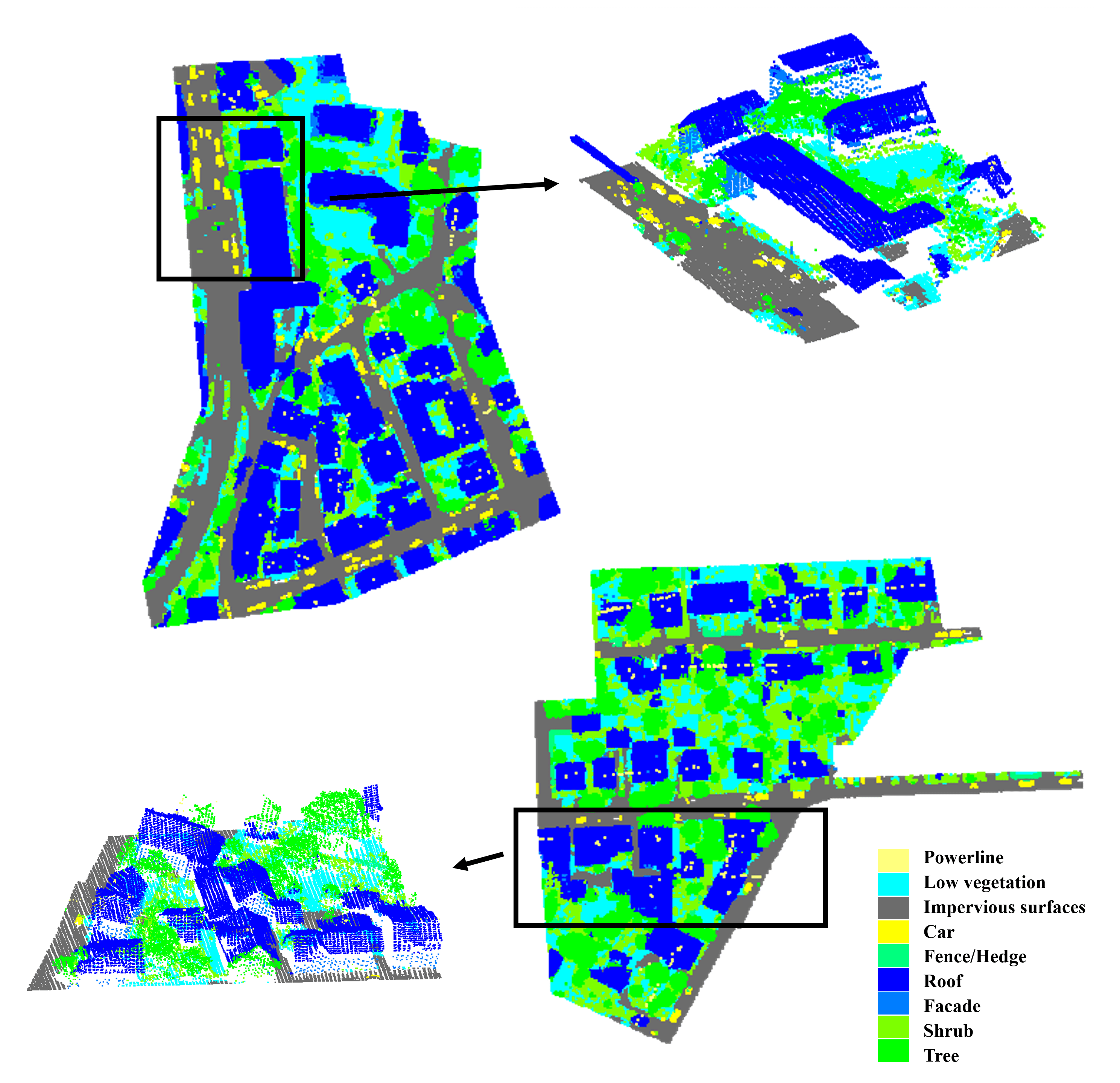}
\caption{The classification results of our proposed D-FCN model on the ISPRS dataset. The black boxes in the figure represent partial enlargements to show details of the classification results. Best viewed in color.
}
\label{fig_Result_overview}
\end{figure*}

After determining the best hyperparameters for the model, we employed Scene I of the ISPRS 3D Semantic Labeling Dataset to train the proposed D-FCN model until convergence. Then, each test block from Scene II and Scene III described in Section \ref{sc_preprocessing} was fed directly into the model without sampling or dropout. The final point cloud classification results and error maps are shown in Figure \ref{fig_Testset_split} and Figure \ref{fig_Result_overview}. As shown in Figure \ref{fig_Testset_split} and Figure \ref{fig_Result_overview}, the proposed D-FCN model successfully generates the correct label predictions for most of the points in the test scenes.

To quantitatively evaluate the classification performance, we calculated the classification confusion matrix, precision, recall and F1 score of each category and list the results in Table \ref{table_confusion_matrix}, where we can see that our proposed model obtained F1 scores above 70\% for six of the categories, including powerline, low vegetation, impervious surfaces, car, roof, and tree.
In addition, our model achieved a reasonable classification performance on the facade category, but its classification performance on the fence/hedge and shrub categories was relatively lower.
As Table \ref{table_confusion_matrix} shows, most of the fence/hedge points were incorrectly classified as shrubs. A primary reason for this result is that the fence/hedge category contains fewer points and presents spatial distribution and topological characteristics similar to the shrub category. These factors cause the model to be insufficiently trained, which prevents the model from being able to differentiate these two categories. Although the powerline and car categories have fewer points in the dataset, they present completely different characteristics from other categories; thus, they result in a better classification performance.

\begin{figure}[h]
\centering
\includegraphics[width=8cm]{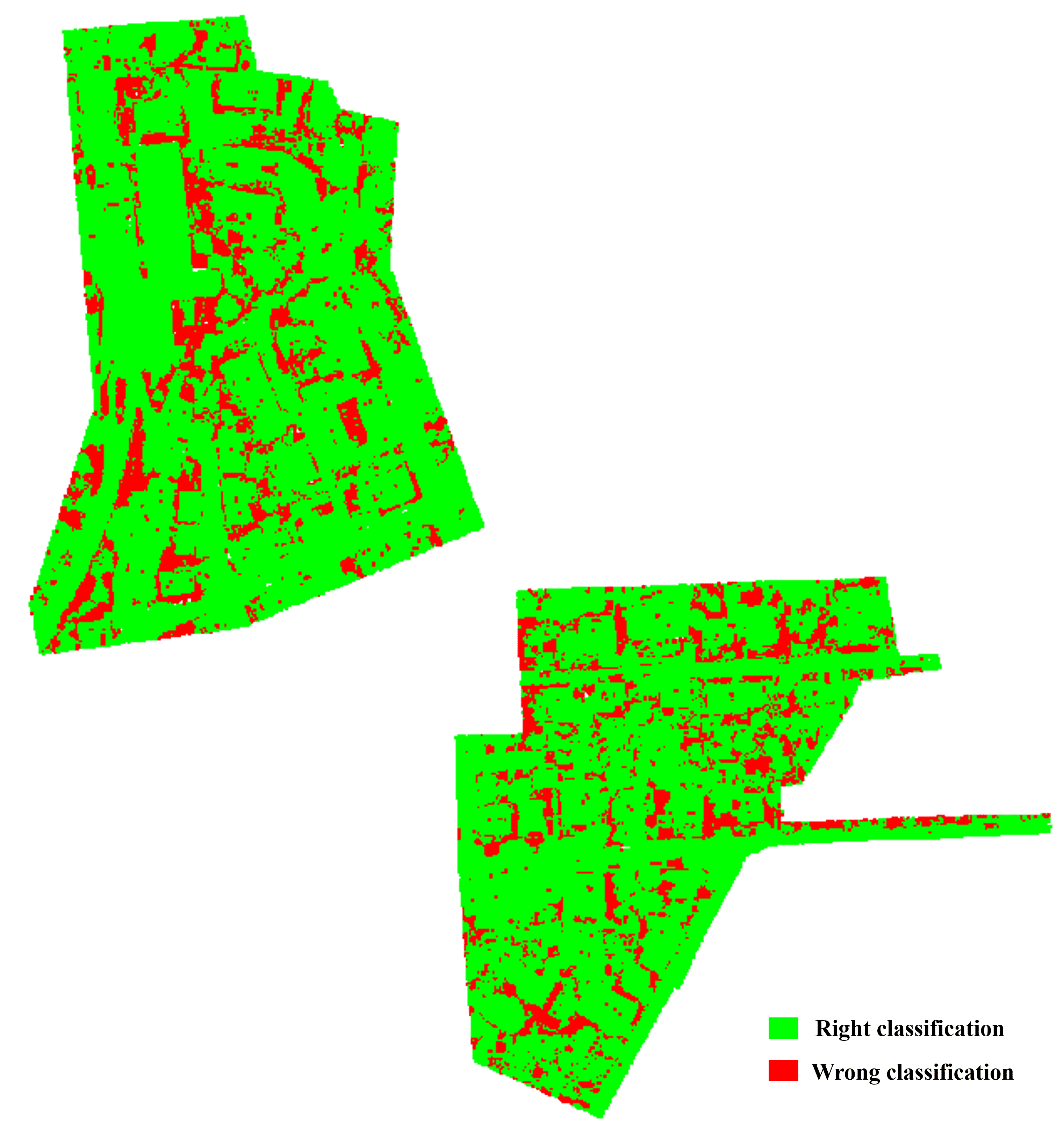}
\caption{The classification error map of our proposed D-FCN model on the ISPRS dataset. The points marked in red and green respectively represent the incorrect and correct classification results. Best viewed in color.
}
\label{fig_Testset_split}
\end{figure}

\subsection{Comparisons with other methods}\label{Experiment_Comparison }

To demonstrate the advantages of the proposed model, we compared it with other models that participated in the ISPRS contest. The top eight models (i.e., UM \citep{horvat2016context}, WhuY2, WhuY3 \citep{yang2017convolutional}, LUH \citep{niemeyer2016hierarchical}, BIJ\_W \citep{wang2018deep}, RIT\_1 \citep{yousefhussien2018multi}, NANJ2 \citep{zhao2018classifying} and WhuY4 \citep{yang2018segmentation}) that achieved the best performance on this benchmark were selected for performance comparisons. The overall accuracy and F1 scores of our model and of all the compared models are listed in Table  \ref{table_avgF1_comparion}. As Table \ref{table_avgF1_comparion} shows, the proposed model achieved superior classification performances than did all the compared methods with regard to average F1 score. In particular, our model achieved substantially higher performances on the powerline, car and facade categories, for which it outperformed the state-of-the-art models by 8.4\%, 3.4\%, 7.3\%, respectively.

However, analyzing the overall classification accuracy, our model failed to achieve the best performance compared to the other state-of-art models. This result occurs mainly because we adopted the average F1 score as the main indicator when determining the initial model parameters rather than overall accuracy. As mentioned in Section \ref{Experiment_Model_Parameters}, many hyper-parameter settings exist that could lead to an overall accuracy above 82.2\% (obtained with the optimal parameters setting aimed at obtaining the best average F1 score); see Table \ref{table_sample_neighbor} and Table\ref{table_balance}. However, because adopting overall accuracy as the main evaluation metric will cause minority categories to be ignored, we focused primarily on the F1 score in this study.

\begin{figure*}[t]
\centering
\includegraphics[width=17cm]{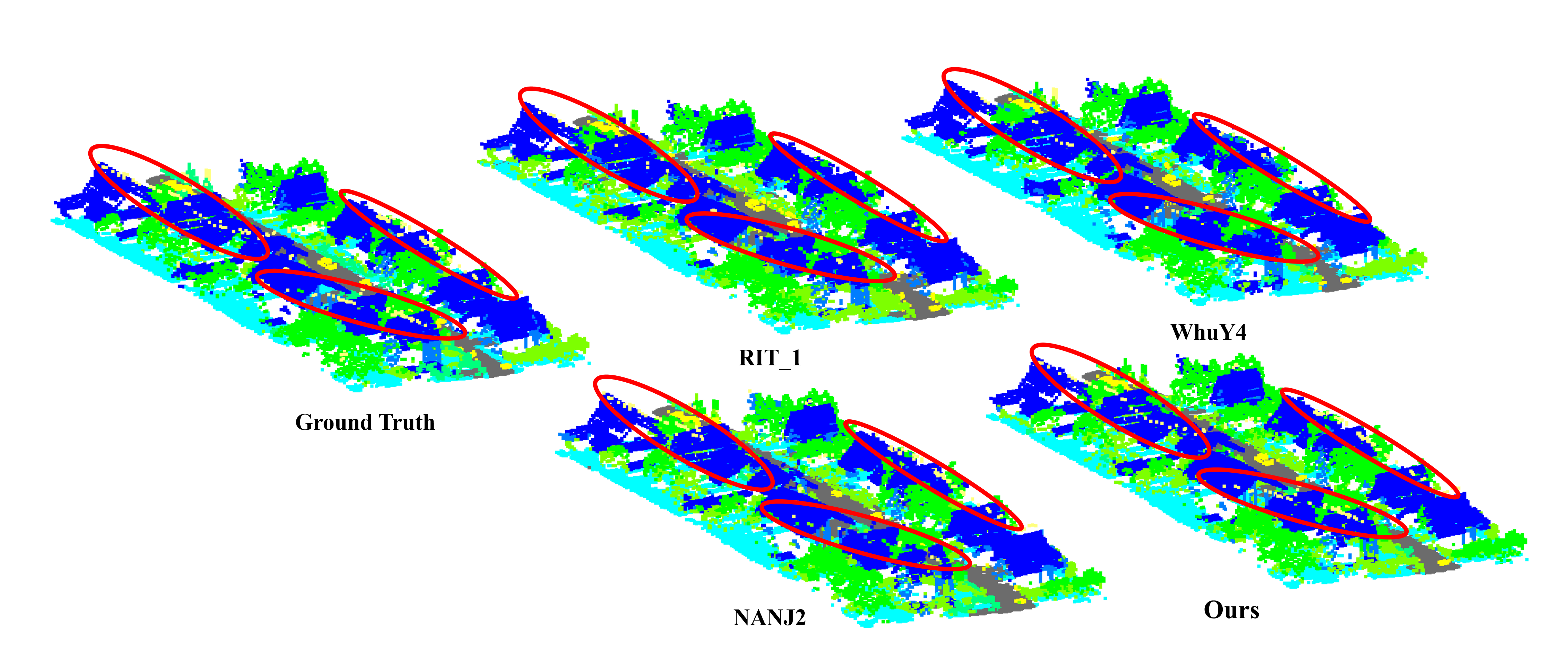}
\caption{The classification results of RIT\_1, WhuY4, NANJ2 and our proposed D-FCN model applied to a selected complicated scene area. The powerline category points are marked in light yellow and highlighted by red circles. This image is best viewed in color.}
\label{fig_Testset_compare}
\end{figure*}

To further compare the classification performance of different models, we selected the top three models from the ISPRS contest (i.e., RIT\_1 \citep{yousefhussien2018multi}, NANJ2 \citep{zhao2018classifying} and WhuY4  \citep{yang2018segmentation}) as comparison models and show their classification results on a selected complicated scene area in Figure \ref{fig_Testset_compare}. Taking the points of the powerline category (marked in light yellow) as an example, our model clearly demonstrates higher classification accuracy compared to the other three models, which misclassify most of the powerline points into other categories. The main reason for this result is that only 600 points exist in the powerline category, accounting for merely 0.1\% of the total 411,722 points in the test set. Therefore, this category is largely ignored by the three compared models.

\begin{table*}
\begin{center}
\caption{Quantitative comparisons between our method and other state-of-art models on the ISPRS benchmark dataset. The numbers in the first nine columns of the table show the F1 scores for each category, and the last two columns show the overall accuracy (OA) and average F1 score (Average F1). The boldface text indicates the model with the best performance.}
\label{table_avgF1_comparion}
\begin{tabular}{l c c c c c c c c c c c}

\hline
Categories & power & low\_veg & imp\_surf & car & fence\_hedge & roof & facade & shrub & tree & OA & Average F1\\
\hline

UM & 0.461 & 0.790 & 0.891 & 0.477 & 0.052 & 0.920 & 0.527 & 0.409 & 0.779 & 0.808 & 0.590 \\
WhuY2 & 0.319 & 0.800 & 0.889 & 0.408 & 0.245 & 0.931 & 0.494 & 0.411 & 0.773 & 0.810 & 0.586 \\ 
WhuY3 & 0.371 & 0.814 & 0.901 & 0.634 & 0.239 & 0.934 & 0.475 & 0.399 & 0.780 & 0.823 & 0.616 \\
LUH & 0.596 & 0.775 & 0.911 & 0.731 & 0.340 & 0.942 & 0.563 & 0.466 & \textbf{0.831} & 0.816 & 0.684 \\
BIJ\_W & 0.138 & 0.785 & 0.905 & 0.564 & 0.363 & 0.922 & 0.532 & 0.433 & 0.784 & 0.815 & 0.603 \\
RIT\_1 & 0.375 & 0.779 & \textbf{0.915} & 0.734 & 0.180 & 0.940 & 0.493 & 0.459 & 0.825 & 0.816 & 0.633 \\
NANJ2 & 0.620 & \textbf{0.888} & 0.912 & 0.667 & 0.407 & 0.936 & 0.426 & \textbf{0.559} & 0.826 & \textbf{0.852} & 0.693 \\
WhuY4 & 0.425 & 0.827 & 0.914 & 0.747 & \textbf{0.537} & \textbf{0.943} & 0.531 & 0.479 & 0.828 & 0.849 & 0.692 \\
\hline
Ours  & \textbf{0.704}  & 0.802  & 0.914  & \textbf{0.781}  & 0.370  & 0.930  & \textbf{0.605}  & 0.460  & 0.794  & 0.822 & \textbf{0.707}\\  

\hline

\end{tabular}
\end{center}
\end{table*}

\section{Conclusions}\label{Conclusion}

In this paper, we proposed a directionally constrained fully convolution neural network (D-FCN) model that can be directly applied to unstructured 3D point sets to classify airborne LiDAR point clouds. Specifically, we proposed a novel directionally constrained point convolution (D-Conv) module to extract the local representative features of 3D point sets from projected 2D receptive fields and a multiscale fully convolutional neural network to enable multiscale point feature learning. Our model can process input point clouds of arbitrary sizes and directly predict the semantic labels for all the input points in an end-to-end manner. To demonstrate the advantages of the proposed model, we conducted experiments on the ISPRS 3D Labeling Benchmark Dataset and compared our model's performance with those of other state-of-art methods. The results show that our model achieved a new state-of-art classification performance for average F1 score despite using only the original 3D coordinates and intensity as inputs. Our D-FCN model outperformed the state-of-the-art methods by 8.4\%, 3.4\%, 7.3\% on the powerline, car, and facade categories, respectively.

\section*{ACKNOWLEDGEMENTS (Optional)}\label{ACKNOWLEDGEMENTS}
We thank the NYU MMVC Lab for providing GPU-equipped servers for these experiments. Additionally, we would like to gratefully acknowledge the ISPRS for providing airborne LiDAR data.%, and we also thankful for the thorough suggestions of the anonymous reviewers and editor.

{%\footnotesize
\begin{spacing}{0.9}% tune the size by altering the parameter
\bibliography{bibliography} % Include your own bibliography (*.bib), style is given in isprs.cls
\end{spacing}
}

% \section*{APPENDIX (Optional)}\label{APPENDIX}

% Any additional supporting data may be appended, provided the paper does not exceed the limits given above.
% \textit{Revised April 2014}

\end{document}